\documentclass[10pt,twocolumn]{article}

\usepackage[
margin=0.75in
]{geometry}

\usepackage{titlesec}
\usepackage{flushend}
\usepackage{enumitem}
\titlespacing{\section}
{0pt}{8pt}{4pt}
\titlespacing{\subsection}
{0pt}{6pt}{3pt}
\usepackage[font=small]{caption}
\usepackage{caption}
\usepackage{times}
\usepackage{graphicx}
\usepackage{amsmath}
\usepackage{amssymb}
\usepackage{booktabs}
\usepackage{multirow}
\usepackage{algorithm}
\usepackage{algorithmic}
\usepackage[numbers]{natbib}
\usepackage{balance}

\usepackage[
    colorlinks=true,
    linkcolor=blue,
    citecolor=blue,
    urlcolor=blue
]{hyperref}

\title{MAMath-Bench: Process-Aware and Robust Evaluation of Multimodal Multi-Agent Mathematical Reasoning}

\author{
Xinping Wang$^{1}$, Jie Ma$^{1,2,*}$, Rui Xing$^{2}$,
Wensong Guo$^{3}$, Xiayi Zhang$^{1}$,\\
Jianhao Deng$^{3}$, Zihao Zhou$^{4}$,
Boqian Yang$^{3}$, Feijuan He$^{5,6}$\\[6pt]
$^{1}$School of Cyber Science and Engineering, Xi'an Jiaotong University\\
$^{2}$MOE KLINNS Lab, Xi'an Jiaotong University\\ 
$^{3}$School of Computer Science and Technology, Xi’an Jiaotong University\\
$^{4}$School of Mathematics and Statistics, Xi'an Jiaotong University\\
$^{5}$Xi'an Jiaotong University City College\\
$^{6}$Engineering Research Center of IoT Intelligent Sensing Interactive Platform, \\
Universities of Shaanxi Province\\[4pt]
$^{*}$Corresponding author\\
\texttt{jiema@xjtu.edu.cn}
}
\date{}

\begin{document}
\maketitle
\setlength{\bibsep}{0pt}

\section*{Abstract}
Mathematical reasoning is a fundamental capability for educational applications of Large Language Models (LLMs), where reliable problem solving requires not only correct answers but also coherent and verifiable reasoning processes. Multi-agent systems (MASs) have shown growing potential in this domain by leveraging collaboration and verification among multiple LLM agents. However, existing evaluations primarily focus on the correctness of final answers in text-only settings, offering limited insight into multimodal reasoning capabilities, the reliability of intermediate reasoning steps, and robustness against reasoning errors. To address these limitations, we introduce MAMath-Bench, a process-aware benchmark designed to evaluate the multimodal mathematical reasoning capabilities of MASs. MAMath-Bench comprises 1,003 problems spanning 11 mathematical categories, multiple difficulty levels, and several modalities—including text, images, and audio. It features fine-grained milestone annotations to evaluate systems across three dimensions: problem-solving correctness, reasoning-process reliability, and robustness to adversarial reasoning-chain perturbations. Our evaluation yields three key findings. First, multi-agent collaboration does not consistently improve final-answer accuracy over strong single LLMs. Second, collaboration can substantially enhance the coverage of intermediate reasoning, but such gains may come at the cost of increased reasoning redundancy. Third, current systems remain vulnerable to adversarial reasoning errors, particularly those involving subtle structural perturbations. The dataset is available at \url{https://github.com/mira-ai-lab/AgentMath}.

\textbf{Keywords:}
multimodal mathematical reasoning,
process-level evaluation,
multi-agent systems

%%%%%%%%%%%%%%%%%%%%%%%%%%%%%%%%%%%%%%%%%%%%%%%%%%%%%%%%%%%%%%%%%%%%%%%%
\section{Introduction}
Recent advances in Large Language Models (LLMs)~\cite{brown2020language, achiam2023GPT, guo2025deepseek} have spurred the development of Multi-Agent Systems (MASs)~\cite{park2023generative, hong2023metaGPT, chen2023agentverse} for complex problem-solving. MASs have been increasingly applied in a variety of real-world domains, including enterprise management~\cite{wang2025bits, xu2025agentic, liu2025workteam}, software programming~\cite{qian2024chatdev, zhang2025knowledge, zhang2026sgagent}, medical diagnosis~\cite{wei2024medco, zhou2025mam, ma2026medla} and smart education~\cite{li2024edumas, mushtaq2025harnessing, xie2026slidebot}. By coordinating multiple agents with specialized roles, such as a planner, a solver, and a verifier, MASs can address complex tasks through sequential workflows~\cite{pmlr-v267-trirat25a} or debate-based interactions~\cite{chan2024chateval}, depending on task requirements. This collaborative design offers a structured framework for decomposing tasks, coordinating intermediate reasoning, and supporting iterative verification. This characteristic is particularly significant in the educational question-answering scenario~\cite{viswanathan2022enhancement, zhang2025eduplanner, bao2026learn}, because students require not only the correct final answers but also a coherent, logical, and verifiable reasoning process~\cite{liu2025knowledge}.

% \mj{To evaluate the performance of MASs in educational question-answering scenarios, a substantial body of research~\cite{lei2024macm, toktobekov2025llm, zhang2025debate4math} has adopted Mathematical Question Answering (MQA) as a standard benchmark, given its stringent demands for structured logical reasoning. However, prevailing evaluation protocols typically restrict problem inputs to unimodal formats~\cite{cobbe2021training, hendrycks2021measuring, glazer2024frontiermath} and assess performance primarily through exact answer matching~\cite{zhou2025reso, zhang2025debate4math, yan2025mathagent}. Despite these advances, we identify three critical limitations in existing evaluation benchmarks.
% }

\begin{figure}[t]
    \centering
    \includegraphics[width=\linewidth]{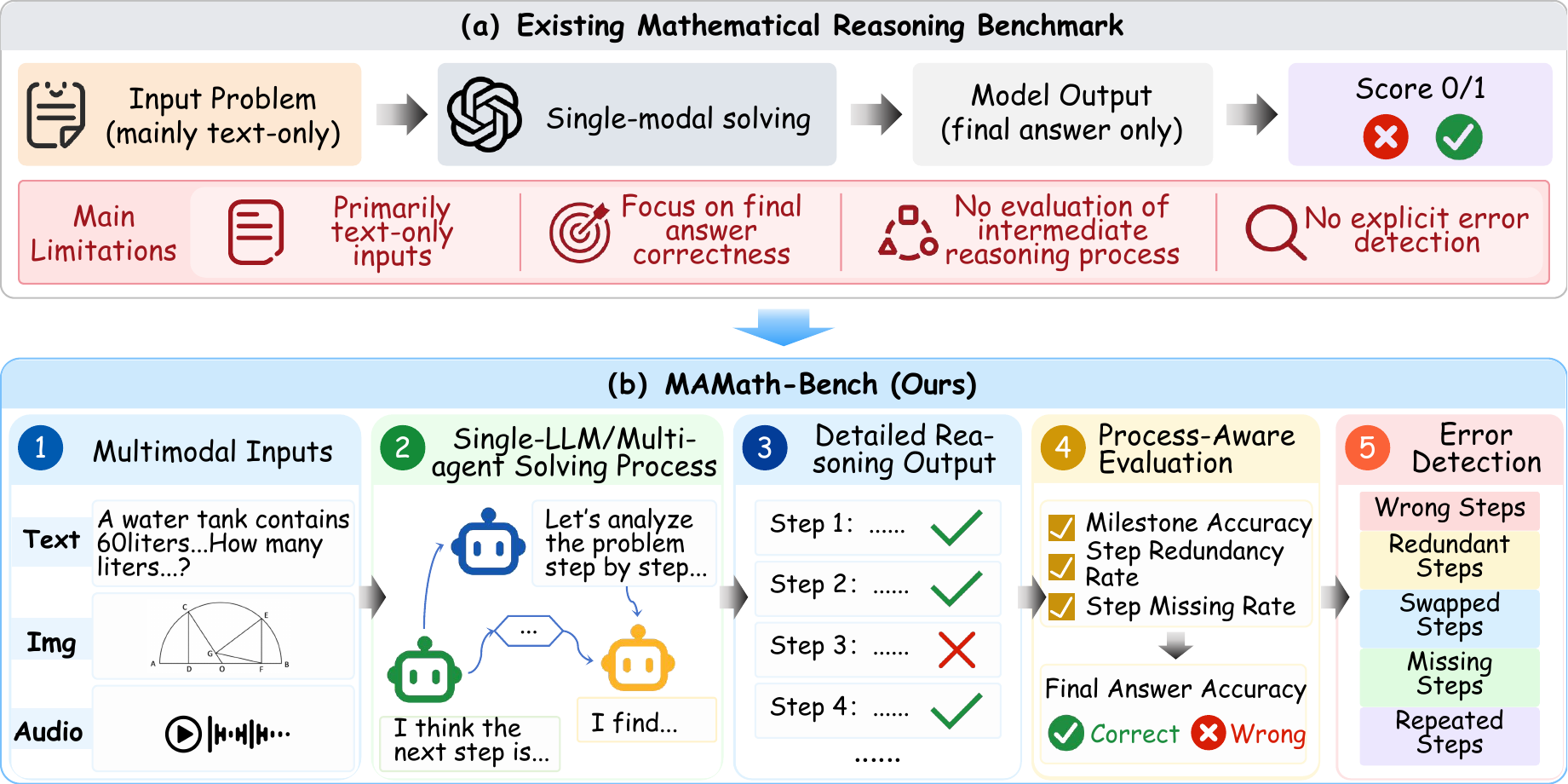}
    \caption{Comparison of a conventional answer-centric evaluation pipeline with the multimodal, process-aware, and error-aware evaluation pipeline of MAMath-Bench.}
    \label{fig:benchmark-comparison}
\end{figure}

To evaluate the performance of MASs in educational question-answering scenarios, a substantial body of research~\cite{wu2023mathchat, lei2024macm} has adopted Mathematical Question Answering (MQA) as a standard benchmark, given its stringent demands for structured logical reasoning. However, the prevailing evaluation protocols typically restrict problem inputs to unimodal formats~\cite{cobbe2021training, hendrycks2021measuring, glazer2024frontiermath} and assess performance primarily through exact response matching~\cite{zhou2025reso, zhang2025debate4math, yan2025mathagent}. However, existing benchmarks do not fully address the evaluation requirements of educational mathematical reasoning, as illustrated in Figure~\ref{fig:benchmark-comparison}. We identify three key limitations.

\textbf{Limited Multimodal Coverage.} In educational settings, mathematical problems are often presented through heterogeneous forms of information, such as textual descriptions, geometric diagrams, symbolic expressions, and spoken instructions. Restricting evaluation to text alone therefore provides limited evidence of whether MASs can integrate and reason over information distributed across different modalities. 

\textbf{Insufficient Process Evaluation.} For educational applications, the reasoning process is itself an important evaluation target, since a correct answer may still be supported by incomplete, redundant, or logically unreliable derivations. Evaluating intermediate reasoning is therefore necessary to distinguish systems that achieve similar answer accuracy but differ substantially in reasoning completeness and reliability.

\textbf{Limited Robustness Assessment.} In educational scenarios, erroneous intermediate steps, misleading solution procedures, or incorrect information from collaborating agents should not be propagated and presented as reliable explanations. Existing benchmarks rarely assess whether MASs can identify and handle such reasoning errors, making it difficult to characterize their reliability under imperfect intermediate conditions. 

To address existing limitations, we introduce MAMath-Bench, which establishes a unified evaluation protocol for systematically assessing and comparing multimodal mathematical reasoning capabilities across different MASs. MAMath-Bench comprises 1,003 carefully curated problems spanning text, visual, and audio modalities. Beyond evaluating final answers, each sample includes milestone-based reasoning annotations, enabling an examination of the reasoning process itself. The benchmark includes \textbf{positive} subsets across three difficulty levels and \textbf{negative} subsets constructed via five types of reasoning perturbations, supporting a three-dimensional evaluation covering problem-solving correctness, collaborative process reliability, and robustness to adversarial reasoning-chain perturbations.

Our main contributions are summarized as follows: 
\begin{itemize}[
    itemsep=0pt,
    parsep=0pt,
    partopsep=0pt
]
\item \textbf{A multimodal benchmark for multi-agent mathematical reasoning.} We propose MAMath-Bench, a benchmark that extends the evaluation of LLM-based MASs from predominantly text-based mathematical reasoning to scenarios involving heterogeneous inputs—including text, vision, and audio. It provides fine-grained annotations, enabling analysis at the levels of both final answers and reasoning processes.

\item \textbf{An evaluation framework balancing reasoning processes and robustness.} Moving beyond the traditional focus on final-answer accuracy, we comprehensively evaluate problem-solving correctness, the reliability of intermediate reasoning, and robustness against adversarial reasoning errors.

\item \textbf{A systematic empirical analysis of multi-agent collaboration mechanisms.} Through extensive comparisons between representative single-agent LLMs and multi-agent frameworks, we find that collaboration does not necessarily yield a universal advantage in final-answer accuracy; instead, it introduces significant trade-offs among reasoning coverage, redundancy, and robustness. These findings reveal a substantial gap between answer correctness and process reliability, offering deep insights into the strengths and limitations of current collaboration mechanisms.
\end{itemize}

\section{Related Work}
\subsection{MQA Datasets}
MQA benchmarks have been extensively studied to evaluate the problem-solving capabilities of LLMs across a range of difficulty levels and reasoning complexities. Among them, GSM8K~\cite{cobbe2021training} is one of the most widely used datasets, consisting of grade-school-level arithmetic word problems that require multi-step reasoning over natural language. To assess robustness, SVAMP~\cite{patel2021nlp} introduces adversarial variations of math word problems designed to reduce reliance on superficial patterns. For more advanced mathematical reasoning, MATH~\cite{hendrycks2021measuring} provides competition-level problems spanning Algebra, Geometry, Number Theory, and Combinatorics, often requiring long and precise chains of symbolic reasoning. Additional challenging benchmarks, such as OlympiadBench~\cite{he2024olympiadbench}, further emphasize rigorous deduction and deeper reasoning abilities. Despite their diversity, these datasets primarily focus on evaluating individual models' reasoning performance, with limited consideration of collaborative or interactive problem-solving processes, which are central to MASs.

\begin{figure*}[t]
  \centering
  \includegraphics[width=\linewidth]{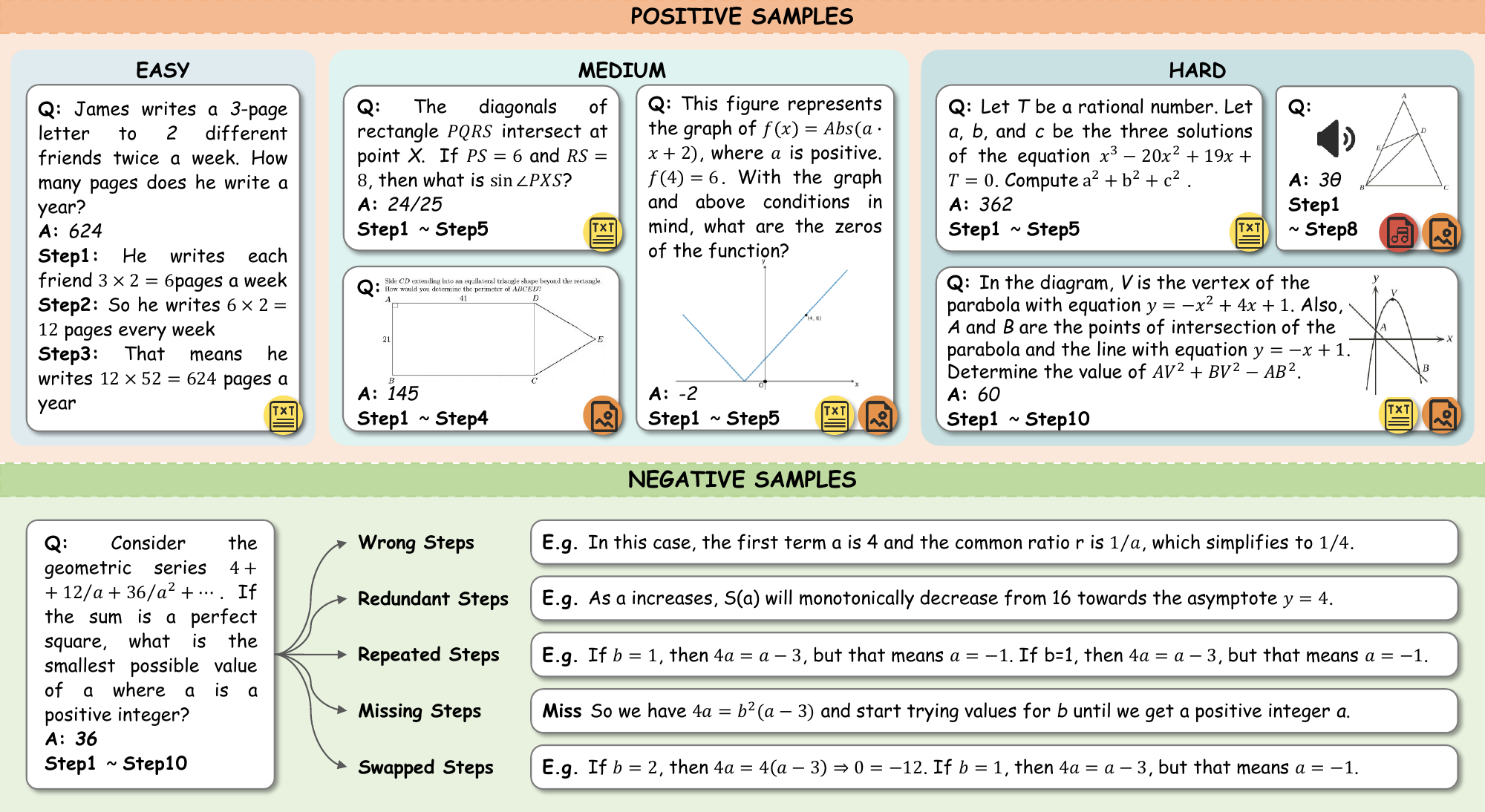}
  \caption{Overview of MAMath-Bench. The positive subset (top) contains mathematical reasoning problems across three difficulty levels and multiple modality configurations. The negative subset (bottom) is constructed by injecting five types of reasoning errors into annotated reasoning chains: wrong, redundant, repeated, missing, and swapped steps.}
  % \Description{High‑level overview diagram for MAMath‑Bench benchmark, illustrating positive multi‑modality samples and five categories of injected reasoning defects for negative samples.}
  \label{fig:benchmark-overview}
\end{figure*}

\subsection{General-Domain Multi-Agent Benchmarks}

Early work on agents primarily focused on evaluating their decision-making~\cite{liu2024agentbench}, planning~\cite{valmeekam2023planbench}, and tool use~\cite{ICLR2024_28e50ee5}, with relatively limited attention to coordination among multiple agents. Recent surveys have systematically examined collaboration mechanisms, failure attribution, and self-evolution in LLM-based multi-agent systems~\cite{Qi2026BeyondII}, highlighting the importance of understanding not only collaborative outcomes but also the underlying interaction and failure processes. More recent benchmarks have increasingly emphasized the quality of collaboration as a dynamic process. For example, BattleAgentBench~\cite{wang2024battleagentbench} decomposes evaluation into progressively challenging stages covering cooperative, competitive, and hybrid interaction settings, explicitly targeting fine-grained collaboration behaviors beyond final task success. Similarly, MultiAgentBench~\cite{zhu2025multiagentbench} introduces diverse scenarios with both aligned and conflicting goals, and evaluates coordination through milestone-based metrics, communication quality, and planning effectiveness.

In addition, recent benchmarks have moved toward more realistic environments. REALM-Bench~\cite{geng2025realm} focuses on real-world planning and scheduling tasks, incorporating metrics such as coordination efficiency, adaptability, resource management, and constraint satisfaction. MAS-Orchestra~\cite{pmlr-v306-ke26c} further studies how factors including reasoning depth, parallelism, and robustness affect multi-agent performance. M3MAD-Bench~\cite{li2026m3mad} standardizes the evaluation of multi-agent debate across domains and modalities, covering both text-only and vision-language settings.

However, these benchmarks are typically designed for general domains such as social interaction, research assistance, and open-domain reasoning, where mathematical reasoning constitutes only a small subset. To the best of our knowledge, there remains a lack of benchmarks specifically designed for multimodal mathematical reasoning in educational settings.
\subsection{Multi-Agent Frameworks for Mathematics}
Motivated by the success of MASs, recent work has explored their application to mathematical reasoning. Debate4Math~\cite{zhang2025debate4math} allows agents to iteratively critique each other's reasoning before reaching consensus, improving performance on challenging benchmarks such as GSM8K~\cite{cobbe2021training} and MATH~\cite{hendrycks2021measuring}. Beyond multi-agent debate, several approaches explore alternative interaction paradigms. For example, MARS\_COLL~\cite{wang2025mars} introduces role-based collaboration, where multiple agents (e.g., author, reviewer, and meta-reviewer) jointly decompose and solve problems, achieving improved accuracy on GSM8K~\cite{cobbe2021training} and other benchmarks containing math questions. AdCo~\cite{liu2025adaptive} investigates hybrid cooperation–competition mechanisms. ECON~\cite{yi2025debate} formulates multi-agent coordination as an incomplete-information game, explicitly modeling the trade-off between accuracy and computational cost.

Other works focus on improving reasoning structure and efficiency. MA-ToT~\cite{haji2024improving} extends Tree-of-Thought reasoning into multi-agent search, while approaches like MACM~\cite{lei2024macm} and ReMA~\cite{NEURIPS2025_b83514e4} incorporate cognitive decomposition and memory mechanisms to enable iterative refinement. Additionally, ILR~\cite{lin2026interactive} introduces interactive learning, allowing agents to improve reasoning strategies over time through feedback. Specialized frameworks such as MathAgent~\cite{yan2025mathagent} further adopt mixture-of-experts designs to handle multimodal mathematical reasoning tasks, while MathChat~\cite{wu2023mathchat} explores dialogue-based collaborative problem solving.

Despite these advances,  a unified evaluation framework that jointly considers multimodal mathematical understanding, intermediate reasoning processes, and robustness remains underexplored. MAMath-Bench is designed to bridge this gap.

\section{MAMath-Bench}

An evaluation dataset for mathematical reasoning in educational scenarios should reflect two characteristics of real learning and assessment. First, students encounter problems with different levels of complexity, requiring an evaluation to distinguish whether reasoning remains effective as the difficulty increases. Second, educational assessment concerns not only the correctness of the final answer, but also the validity of the reasoning process, since intermediate errors can reveal misunderstandings even when the final answer is correct. These considerations motivate the two-dimensional design of MAMath-Bench, which is organized along the dimensions of problem difficulty and reasoning quality, as illustrated in Figure~\ref{fig:benchmark-overview}. The difficulty dimension comprises easy, medium, and hard problems, while the reasoning-quality dimension introduces positive and negative samples. Positive samples cover text, symbolic expressions, geometric diagrams, and audiovisual content with complete and correct reasoning. Negative samples inject five representative reasoning flaws—wrong, redundant, repeated, missing, and swapped steps—so that the benchmark can assess whether agents can recognize and maintain reliable reasoning processes in educational mathematics tasks.

\subsection{Annotation}
We employ a four-stage expert annotation pipeline covering data preparation, milestone annotation, adversarial perturbation construction, and quality verification. All annotations are conducted by annotators with mathematical training and subsequently subjected to cross-review and expert arbitration.

\subsubsection{Raw Data Collection and Difficulty Classification}
To construct a representative and well-structured evaluation set for educational mathematical reasoning, we develop the benchmark data through three stages: source collection, quality filtering, and difficulty organization. 

\textbf{Dataset Sources.} To accurately capture the reasoning performance of large models at different levels of cognitive complexity, we collect mathematical problems from a diverse set of established benchmarks and open-source repositories. Specifically, GSM8K~\cite{cobbe2021training} and SVAMP~\cite{patel2021nlp} are used to provide relatively short-chain arithmetic and word-problem reasoning; MAVIS-Geometry and MAVIS-Function contribute visually grounded mathematical problems involving geometric structures and functional relationships; and more challenging competition-level problems are drawn from the Level-5 subset of Competition Math and OlympiadBench~\cite{he2024olympiadbench}. Together, these sources cover 11 major mathematical problem types and provide a broad range of reasoning depths and modality configurations.

\textbf{Dataset Filtering.} We adopt a dual filtering mechanism of manual and expert rules to eliminate \textit{redundant problems} that are computationally cumbersome but logically simple, and prioritize problems that require substantial spatial reasoning, number-theoretic deduction, and combinatorial reasoning. In this way, the selected samples are better suited for evaluating the upper limits of agents' reasoning capability.

\textbf{Dataset Organization.} We further organize the samples into a progressive difficulty hierarchy according to reasoning-chain length, modality composition, and knowledge-domain complexity. Based on these criteria, the benchmark is divided into easy, medium, and hard levels to better reflect the emphasis on conceptual understanding and multi-step reasoning in mathematics education. 
\begin{itemize}[
    itemsep=0pt,
    parsep=0pt,
    partopsep=0pt
]
    \item \textbf{Easy-level:} These samples involve 1–3 milestones and consist entirely of text-based data.
    \item \textbf{Medium-level:} They involve 4-6 milestones, comprising 100 text-based, 100 image-based entries and 100 text-image entries.
    \item \textbf{Hard-level:} They involve 6-20 milestones, comprising 100 text-based, 100 image-based, and 100 audiovisual entries. 
\end{itemize}
\subsubsection{Milestone Annotation}

We represent each reference reasoning process as an ordered sequence of milestones $M = \{m_1, m_2, \dots, m_n\}$. Each milestone corresponds to a self-contained reasoning unit consisting of the required conditions, the applied rule or theorem, and the resulting intermediate conclusion. This design ensures that each milestone is sufficiently self-contained for independent verification while preserving the logical dependencies between successive reasoning stages.
\begin{figure}[t]
  \centering
  \includegraphics[width=\columnwidth]{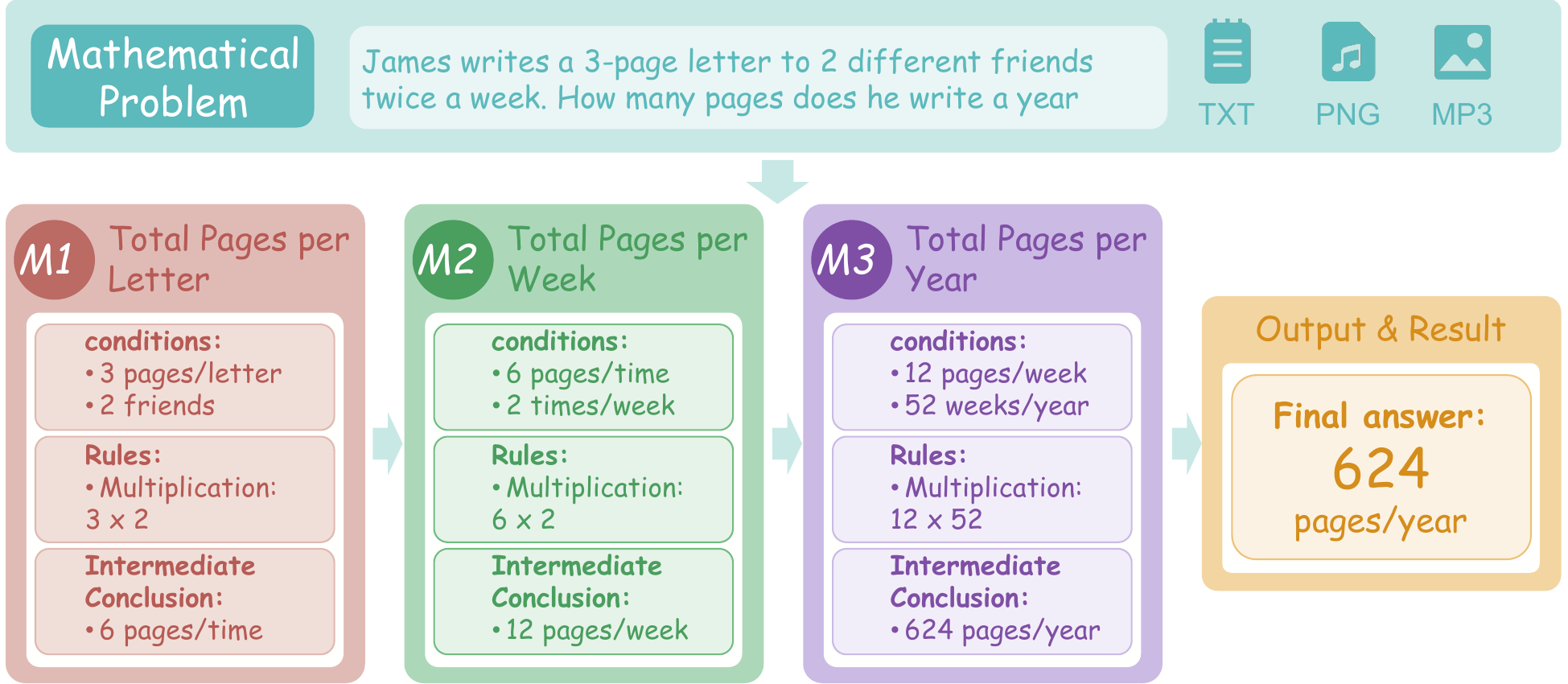}
  \caption{Example of milestone decomposition for a mathematical reasoning problem.}
  \label{fig:milestone}
\end{figure}

To clarify the logical boundaries and core content of each milestone, the annotation team performed fine-grained milestone decomposition on all 800 positive samples. An example of milestone decomposition is shown in Figure~\ref{fig:milestone}.

Furthermore, some mathematical problems admit multiple valid reasoning strategies, making evaluation against a single reference trajectory potentially restrictive. For problems identified during annotation as admitting distinct valid solution paths, we therefore provide multiple milestone sequences. In total, 63 samples are annotated with multiple solution paths. 

\subsubsection{Adversarial Negative Sample Construction}

To systematically evaluate whether MASs can identify errors arising within otherwise plausible reasoning processes, we construct negative samples by applying controlled logical perturbations to valid reasoning chains. Rather than introducing arbitrary noise, these perturbations preserve the original problem context and overall reasoning structure while modifying specific properties of the reasoning process, allowing system failures to be attributed to identifiable reasoning defects. Following the evaluation criteria of ReasonEval~\cite{xia2025evaluating}, we consider five representative types of reasoning defects. All negative samples follow the \textit{single-defect-injection} principle to enable category-wise analysis of error detection capability.

Specifically, the five perturbation types are constructed as follows. 
\begin{itemize}[
    itemsep=0pt,
    parsep=0pt,
    partopsep=0pt
]
    \item \textbf{Wrong Steps} are generated by introducing calculation errors into valid reasoning milestones, such as modifying arithmetic operations or numerical results while keeping the surrounding reasoning context unchanged. 
    \item \textbf{Redundant Steps} are created by inserting unnecessary reasoning milestones that do not contribute additional information or logical progress to the solution, while remaining mathematically plausible. 
    \item \textbf{Repeated Steps} are generated by duplicating existing reasoning content, including both completely repeated milestones and partially repeated consecutive steps, leading to unnecessary repetition in the reasoning chain. 
    \item \textbf{Missing Steps} are constructed by removing essential milestones from the original reasoning trajectory, resulting in incomplete logical transitions between intermediate conclusions. 
    \item \textbf{Swapped Steps} are generated by exchanging the order of two dependent milestones in the original gold reasoning chain. Such perturbations preserve the involved reasoning components but introduce inconsistencies in the dependency order, causing subsequent numerical computations or logical deductions to become misaligned.
\end{itemize}
\subsubsection{Quality Control and Consistency Assessment}

To ensure annotation quality and academic rigor, we design a two-round cross-validation mechanism to strictly control the annotation process and results, ensuring consistency and accuracy. The specific process is as follows:

\begin{itemize}[
    itemsep=0pt,
    parsep=0pt,
    partopsep=0pt
]
    \item \textbf{First-Round Blind Review:} Eight graduate students majoring in mathematics and computer science are paired up to conduct blind reviews of milestone decompositions and negative sample defect annotations for all samples. They are unaware of the sample source and the research hypotheses. Samples with inconsistent annotations are marked as disputed samples.
    \item \textbf{Second-Round Arbitration:} An arbitration panel composed of two computer science professors reviews each disputed sample, clarifies annotation standards, and reaches a consensus to ensure the rigor of the annotation results.
\end{itemize}

\subsection{Dataset Statistics}
\label{sec:dataset_statistic}
\begin{figure}[t]
  \centering
  % 左侧：表格 minipage [t]顶部对齐
  \begin{minipage}[c]{0.47\columnwidth}
    \captionof{table}{Overall Statistics of MAMath-Bench.}
    \label{tab:dataset_stats}
    {\small
    \setlength{\tabcolsep}{2pt} % 默认6pt，缩小列间距
    \centering
    \begin{tabular}{p{2.0cm}r}
      \toprule
      \textbf{Item} & \textbf{Number} \\
      \midrule
      Total samples                  & 1,003 \\
      \midrule
      Positive         & 800  \\
      Text/Image           & 400, 100  \\
      Text-Image             & 200  \\
      Audio-Image           & 100  \\
      \midrule
      Negative         & 203  \\
      Wrong/Missing        & 83, 30 \\
      Repeated/Swapped  &30, 30\\
      Redundant     & 30 \\
      \midrule
      easy/medium     & 200, 300\\
      hard              & 300\\  
      \bottomrule
    \end{tabular}
    }
  \end{minipage}%
  \hfill
  % 右侧：图片 minipage [t]顶部对齐
  \begin{minipage}[c]{0.50\columnwidth}
    \includegraphics[width=\linewidth]{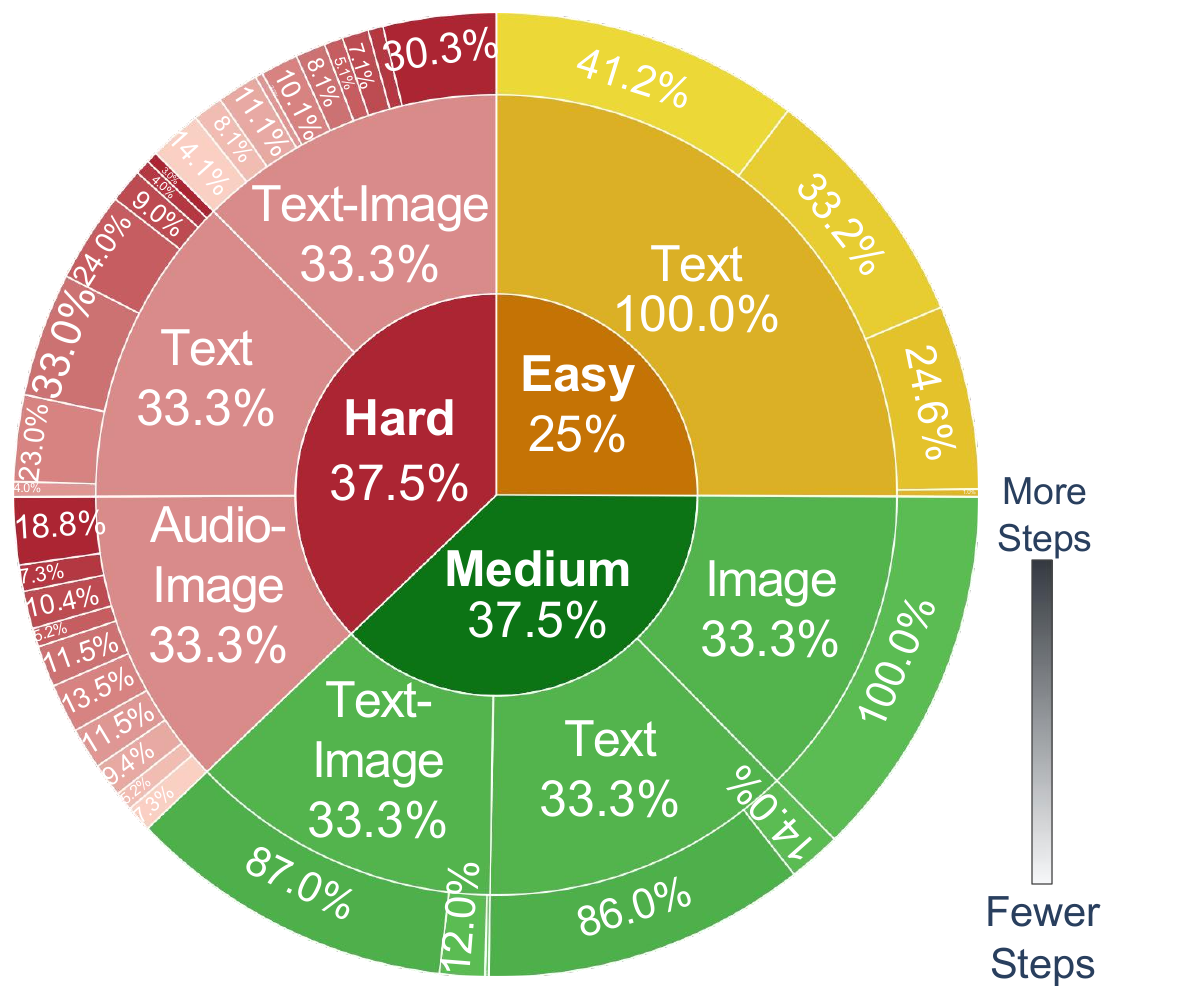}
    % \Description{Bar chart showing the difficulty stratification of reasoning samples.}
    \captionof{figure}{The difficulty stratification of reasoning samples.}
    \centering
    \label{fig:pdf_figure}
  \end{minipage}%
\end{figure}

MAMath-Bench divides 1,003 samples into three difficulty levels based on the length of the reasoning chain and the perceptual complexity. As the difficulty increases, the core statistical information is shown in Figure~\ref{fig:pdf_figure} and Table~\ref{tab:dataset_stats}.

The distribution of milestones reflects distinct levels of difficulty. Easy-level samples involve 1–3 milestones and consist entirely of text-based data; Medium-level samples involve 4-6 milestones, comprising 100 text-based, 100 image-based entries and 100 text-image entries; and Hard-level samples involve 6-20 milestones, comprising 100 text-based, 100 image-based, and 100 audiovisual entries. This distribution directly facilitates the evaluation of MASs' capabilities regarding task division and collaboration.

A total of 203 adversarial negative samples are constructed for the adversarial dataset. Serving as a core component of the testing, the Wrong Steps defect category aims to determine whether a multi-agent system relies merely on result matching for superficial verification or possesses the genuine capability to conduct a deep logical audit of the reasoning chain.

\section{Experiments}

To systematically evaluate multimodal mathematical reasoning capabilities, we structure our experiments around three core research questions:
\begin{itemize}[
    itemsep=0pt,
    parsep=0pt,
    partopsep=0pt
]
    \item \textbf{RQ1:} How accurately can current single-LLM and multiagent systems solve multimodal mathematical problems?
    \item \textbf{RQ2:} How reliable are their intermediate reasoning steps?
    \item \textbf{RQ3:} How robust are these systems to adversarial perturbations in the reasoning chain?
\end{itemize}

\subsection{Experimental Setup}
\subsubsection{Datasets}
All models are evaluated on MAMath-Bench under a unified
evaluation protocol. The positive split is used to assess
final-answer correctness and intermediate reasoning quality. For negative samples, the problem and its perturbed reasoning chain are provided as input to evaluate the ability to detect reasoning anomalies. Performance is reported both overall and by error type.

\subsubsection{Non-Collaborative Baselines and Collaborative MASs}
To comprehensively assess mathematical reasoning capabilities in educational scenarios and examine the benefits of collaborative reasoning, we evaluate both representative single LLMs, which serve as non-collaborative reference baselines, and multi-agent reasoning frameworks. Based on their relevance to mathematical reasoning and representation of distinct collaboration paradigms, we selected four MASs, including MACM~\cite{lei2024macm}, MARS\_COLL~\cite{wang2025mars}, Debate4MATH~\cite{zhang2025debate4math}, and AdCo~\cite{liu2025adaptive}. 

For all multi-agent baselines, we follow their official open-source implementations and preserve the original agent roles, interaction protocols, and system prompts whenever applicable. To ensure a consistent evaluation setting, only the task-specific descriptions required to present MAMath-Bench problems are standardized across systems, while the remaining prompts and collaboration mechanisms are kept consistent with their original implementations. In addition, all evaluated MASs use GPT-4.1 as their backbone model.

\begin{table}[ht]
  \centering
  \caption{Modalities supported by evaluated models.}
  \label{tab:model_modality}
  \small
  \begin{tabular}{lccc}
    \toprule
    \textbf{Model} & \textbf{Text} & \textbf{Image} & \textbf{Audio} \\
    \midrule
    \multicolumn{4}{l}{\textbf{Single LLM}} \\
    Qwen2.5-Math-7B-Instruct & \checkmark & $-$ & $-$ \\
    Qwen2.5-Math-72B-Instruct & \checkmark & $-$ & $-$ \\
    DeepSeek-V3.2-Thinking & \checkmark & $-$ & $-$ \\
    DeepSeek-V4-Pro & \checkmark & $-$ & $-$ \\
    Qwen3-VL-8B-Thinking & \checkmark & \checkmark & $-$ \\
    Qwen3-VL-32B-Thinking & \checkmark & \checkmark & $-$ \\
    Qwen3-Omni-30B-A3B-Thinking & \checkmark & \checkmark & \checkmark \\
    GPT-4.1 & \checkmark & \checkmark & $-$ \\
    GPT-5.4 & \checkmark & \checkmark & $-$ \\
    Gemma-4-E4B & \checkmark & \checkmark & \checkmark \\
    \midrule
    \multicolumn{4}{l}{\textbf{Multi-Agent Framework}} \\
    MACM & \checkmark & \checkmark & $-$ \\
    MARS\_COLL & \checkmark & \checkmark & $-$ \\
    Debate4MATH & \checkmark & \checkmark & $-$ \\
    AdCo & \checkmark & \checkmark & $-$ \\
    \bottomrule
  \end{tabular}
\end{table}

Importantly, the models evaluated differ in their support for the native modality. We therefore explicitly record the modality accessibility of each evaluated system and avoid treating unsupported modalities as failed predictions, as shown in Table~\ref{tab:model_modality}.

\subsubsection{Metrics}
We adopt four primary metrics to capture correctness, process quality, and efficiency.
\begin{itemize}[
    itemsep=0pt,
    parsep=0pt,
    partopsep=0pt
]
    \item \textbf{Final-Answer Accuracy (FAA)}: The exact-match accuracy of the final numerical or symbolic answer. It is defined as
\begin{equation}
\mathrm{FAA}=\frac{N_{\mathrm{correct}}}{N},
\end{equation}
where $N_{\mathrm{correct}}$ is the number of correctly answered questions, and $N$ is the total number of questions evaluated.

\item \textbf{Milestone Accuracy (MA)}: The proportion of correctly identified intermediate reasoning milestones. Each reference milestone $g$ is aligned with the predicted reasoning unit $p$ achieving the highest composite similarity, calculated from four components:

(1) \textit{Semantic similarity} combines word-level and character-level TF-IDF cosine similarities as $S_{\mathrm{sem}}=0.65\cos(\mathbf{w}_p,\mathbf{w}_g)+0.35\cos(\mathbf{c}_p,\mathbf{c}_g)$.

(2) \textit{Equation consistency} is defined as $S_{\mathrm{eq}}=\mathbb{I}[\exists\,e_p\in\mathcal{E}(p),e_g\in\mathcal{E}(g):e_p\equiv e_g]$, where $\mathcal{E}$ denotes extracted mathematical expressions, and equivalence is assessed through normalized comparison, numerical equivalence, or symbolic simplification.

(3) \textit{Numerical overlap} measures the proportion of shared
numerical values, defined as
$S_{\mathrm{num}}=|\mathcal{N}(p)\cap\mathcal{N}(g)|/|\mathcal{N}(g)|$,
where $\mathcal{N}$ denotes extracted numerical values.
The score is set to zero if $\mathcal{N}(g)$ is empty.

(4) \textit{Keyword overlap} uses Jaccard similarity, $S_{\mathrm{key}}=|\mathcal{K}(p)\cap\mathcal{K}(g)|/|\mathcal{K}(p)\cup\mathcal{K}(g)|$, where $\mathcal{K}$ denotes normalized token sets.

The overall matching score is
\begin{equation}
S(p,g)=0.55S_{\mathrm{sem}}+0.25S_{\mathrm{eq}}
+0.10S_{\mathrm{num}}+0.10S_{\mathrm{key}}.
\end{equation}
The weights are heuristically assigned to prioritize semantic
similarity (0.55) and mathematical expression agreement (0.25),
which capture the primary content of reasoning milestones.
Numerical consistency (0.10) and keyword overlap (0.10)
serve as complementary signals to refine the matching process. A reference milestone is matched if its best-aligned predicted unit satisfies $S\geq0.4$, $S_{\mathrm{eq}}=1$, or $S_{\mathrm{sem}}\geq0.4\land(S_{\mathrm{num}}\geq0.5\lor S_{\mathrm{key}}\geq0.2)$. The threshold of 0.4 is used to accommodate variations in reasoning-step expressions while maintaining a minimum level of similarity for milestone matching. These criteria remain fixed across all evaluated models.

Finally, MA is defined as
\begin{equation}
\mathrm{MA}=\frac{N_{\mathrm{matched}}}{N_{\mathrm{gold}}},
\end{equation}
where $N_{\mathrm{matched}}$ and $N_{\mathrm{gold}}$ denote the numbers of matched and total reference milestones, respectively.

    \item \textbf{Step Redundancy Rate (SRR)}: The percentage of excess generated reasoning steps beyond the required reasoning coverage. It is defined as
\begin{equation}
\mathrm{SRR}=\frac{N_{\mathrm{redundant}}}{N_{\mathrm{pred}}},
\end{equation}
where $N_{\mathrm{pred}}$ is the total number of generated reasoning steps and $N_{\mathrm{redundant}}$ is computed as $N_{\mathrm{redundant}}=\max(0, N_{\mathrm{pred}}-\max(N_{\mathrm{matched\_pred}},N_{\mathrm{gold}}))$. Here, $N_{\mathrm{matched\_pred}}$ represents the number of generated steps aligned with reference milestones, and $N_{\mathrm{gold}}$ denotes the number of reference milestones. A higher SRR indicates that the model introduces more unnecessary reasoning overhead beyond the required reasoning coverage.
    \item \textbf{Step Missing Rate (SMR)}: The percentage of essential steps omitted. It is defined as
\begin{equation}
\mathrm{SMR}=\frac{N_{\mathrm{gold}}-N_{\mathrm{matched}}}{N_{\mathrm{gold}}}.
\end{equation}
Thus, SMR is complementary to MA and directly reflects omissions in the reasoning process.
\end{itemize}

Together, these metrics operationalize process-level reliability from complementary perspectives: MA measures the coverage of reference reasoning milestones, SMR reflects milestone omission, and SRR captures excessive or unmatched reasoning steps. Therefore, a reasoning trace with high MA and low SMR/SRR is considered more reliable in terms of alignment, completeness, and conciseness with respect to the reference reasoning process.

\begin{table}[h]
\centering
\caption{Performance comparison between single LLMs and multi-agent frameworks (\%).
\textit{Abbreviations: Qwen2.5-Math-7B-Instruct (Qwen2.5-Math-7B), Qwen2.5-Math-72B-Instruct (Qwen2.5-Math-72B), Qwen3-VL-8B-Thinking (Qwen3-VL-8B), Qwen3-VL-32B-Thinking (Qwen3-VL-32B), Qwen3-Omni-30B-A3B-Thinking (Qwen3-Omni-30B), DeepSeek-v3.2-thinking (DeepSeek-v3.2), DeepSeek-V4-Pro (DeepSeek-V4-Pro), Gemma-4-E4B (Gemma-4-E4B).}}
\label{tab:text_result}
\setlength{\tabcolsep}{3pt}{
\begin{tabular}{lcccc}
\toprule
\textbf{Source} & \textbf{FAA $\uparrow$} & \textbf{MA $\uparrow$} & \textbf{SRR $\downarrow$} & \textbf{SMR $\downarrow$} \\
\midrule
\multicolumn{5}{l}{\textit{Single LLM}} \\
\midrule
Qwen2.5-Math-7B              & 72.04 & 54.28 & 40.96 & 45.72 \\
Qwen2.5-Math-72B             & 75.38 & 51.80 & 38.73 & 48.20 \\
Qwen3-VL-8B                  & 52.01 & 66.44 & 72.04 & 33.56 \\
Qwen3-VL-32B                 & 52.51 & 65.75 & 66.75 & 34.25 \\
Qwen3-Omni-30B               & 41.85 & \textbf{72.50} & 69.62 & \textbf{27.50} \\
GPT-4.1                       & 75.94 & 43.64 & 44.69 & 56.36 \\
GPT-5.4                       & \textbf{79.65} & 45.85 & \textbf{38.33} & 54.15 \\
DeepSeek-v3.2                 & 64.57 & 46.46 & 45.57 & 53.54 \\
DeepSeek-V4-Pro               & 68.67 & 30.60 & 39.36 & 69.40 \\
Gemma-4-E4B                   & 56.64 & 36.71 & 43.58 & 63.29 \\
\midrule
\multicolumn{5}{l}{\textit{Multi-Agent Framework}} \\
\midrule
Debate4MATH                   & 66.50 & 42.83 & 58.80 & 57.17 \\
AdCo                          & 57.04 & 50.59 & \textbf{20.88} & 49.41 \\
MARS\_COLL                    & 68.50 & \textbf{85.80} & 92.17 & \textbf{14.20} \\
MACM                          & 58.76 & 58.38 & 70.50 & 41.62 \\
\bottomrule
\end{tabular}
}    
\end{table}

For negative samples, we additionally report \textbf{Detection Accuracy} per perturbation type. A model is considered correct only if it identifies the specific location and category of the error. The five types of perturbations are wrong steps, redundant steps, repeated steps, missing steps, and swapped steps.
\subsection{Multimodal Mathematical Reasoning Performance}
\subsubsection{Overall Comparison}
Table~\ref{tab:text_result} presents the performance of all evaluated systems on text-based mathematical reasoning tasks. 
Among non-collaborative baseline models, GPT-5.4 achieved the highest FAA. In contrast, MARS\_COLL, the best-performing multi-agent framework, achieved an answer accuracy of 68.5\%, which is lower than the accuracy achieved by its backbone model, GPT-4.1. This indicates that existing multi-agent collaboration methods do not always outperform powerful LLMs in terms of FAA. 

However, the picture changes when the reasoning process is considered. MARS\_COLL achieved a substantially higher MA (85.8\%), surpassing all evaluated LLMs. This result shows MARS\_COLL's strength in covering intermediate reasoning steps, while also revealing its limitations in translating comprehensive reasoning into accurate and efficient problem solving. 

Process quality metrics further reveal distinct reasoning behaviors among collaborative systems. MARS\_COLL achieves more comprehensive coverage of intermediate reasoning steps but introduces substantial redundancy, whereas AdCo generates more concise reasoning chains at the expense of reasoning completeness and final-answer accuracy. These findings highlight a trade-off between reasoning completeness and efficiency across different multi-agent collaboration mechanisms.

Overall, the results demonstrate that evaluating multi-agent mathematical reasoning based solely on FAA is insufficient. Different systems exhibit distinct trade-offs between answer correctness, reasoning completeness, and reasoning efficiency.

\subsubsection{Modality-Wise Performance}

\begin{figure}[t]
  \centering
  \includegraphics[width=\columnwidth]{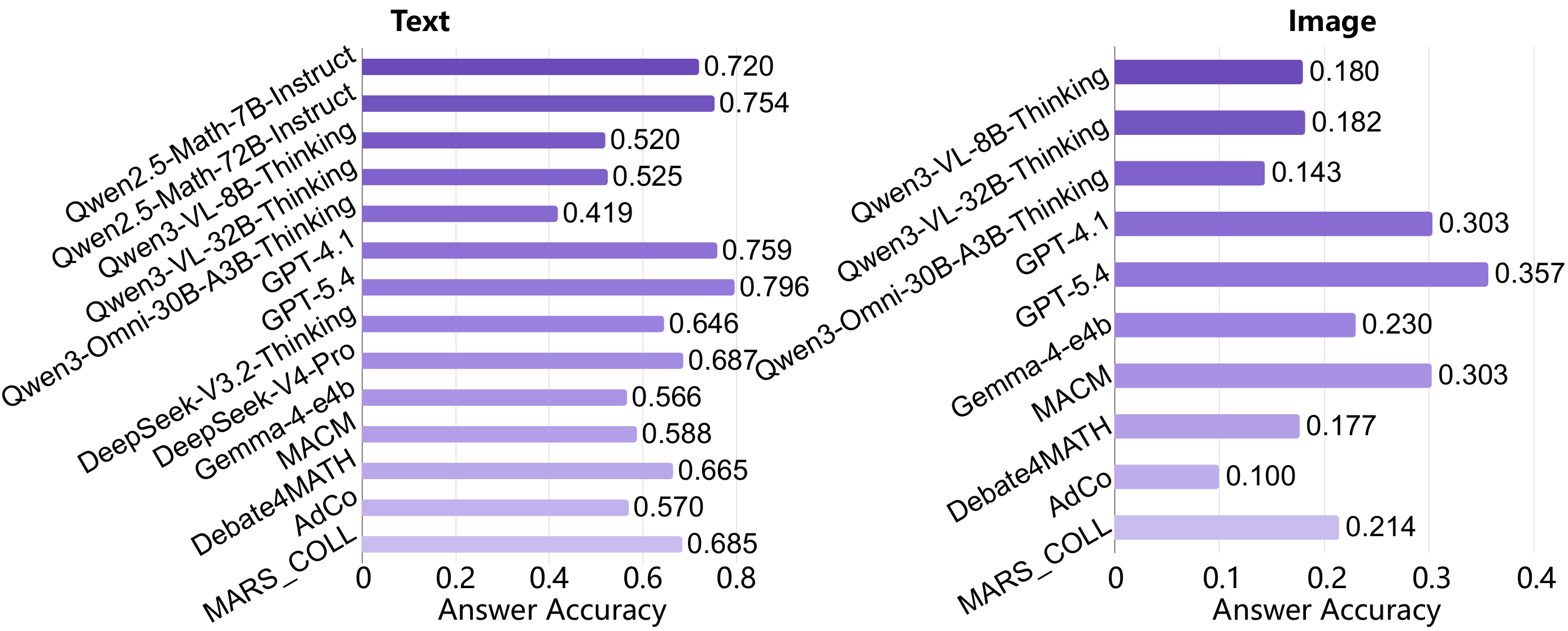}
  \caption{FAA on text- and image-modality mathematical reasoning tasks.}
  % \Description{Accuracy of mathematical reasoning across different modalities.}
  \label{fig:Modality-wise Answer Accuracy}
\end{figure}

Figure~\ref{fig:Modality-wise Answer Accuracy} compares the FAA under text-input and image-input settings. Notably, comparisons between models are limited to the subsets of modalities each is capable of handling.
For text inputs, several strong models achieved accuracy rates exceeding 70\%. GPT-5.4 performed best with an accuracy of 79.65. Among MASs, MARS\_COLL achieved 68.5\%.

In contrast, accuracy on image-based math problems dropped sharply. The top-performing model was GPT-5.4 (35.7\%). Accuracy rates for other systems were generally below 25\%. This widespread performance decline indicates that mathematical reasoning capabilities are strongly influenced by the perceptual demands associated with visual input. Systems must correctly extract mathematical symbols, geometric relationships, or functional structures from images before symbolic reasoning can take place. Consequently, errors arising during the visual perception stage can accumulate and propagate throughout the reasoning process, leading to lower final results.

Among the evaluated multi-agent frameworks, MACM performed best on image-based problems, achieving an accuracy of 30.3\%, on par with GPT-4.1 and substantially higher than MARS\_COLL, Debate4MATH, and AdCo. This result suggests that the effectiveness of multi-agent collaboration under multimodal inputs may be associated with how information is decomposed and exchanged among agents. MACM explicitly extracts and aggregates task-relevant conditions before solving the problem, helping organize vision-based mathematical constraints. However, its performance still dropped significantly from the text modality (58.8\%) to the image modality (30.3\%), indicating that collaborative reasoning cannot fully compensate for errors introduced during visual perception.

\subsubsection{Performance across Difficulty Levels}

To isolate reasoning difficulty from modality effects, we conduct the difficulty-wise analysis exclusively on text-based problems. Image- and audio-based problems occur only in the Medium and Hard subsets and are therefore excluded from this analysis.
\begin{figure}[t]
  \centering
  \includegraphics[width=\columnwidth]{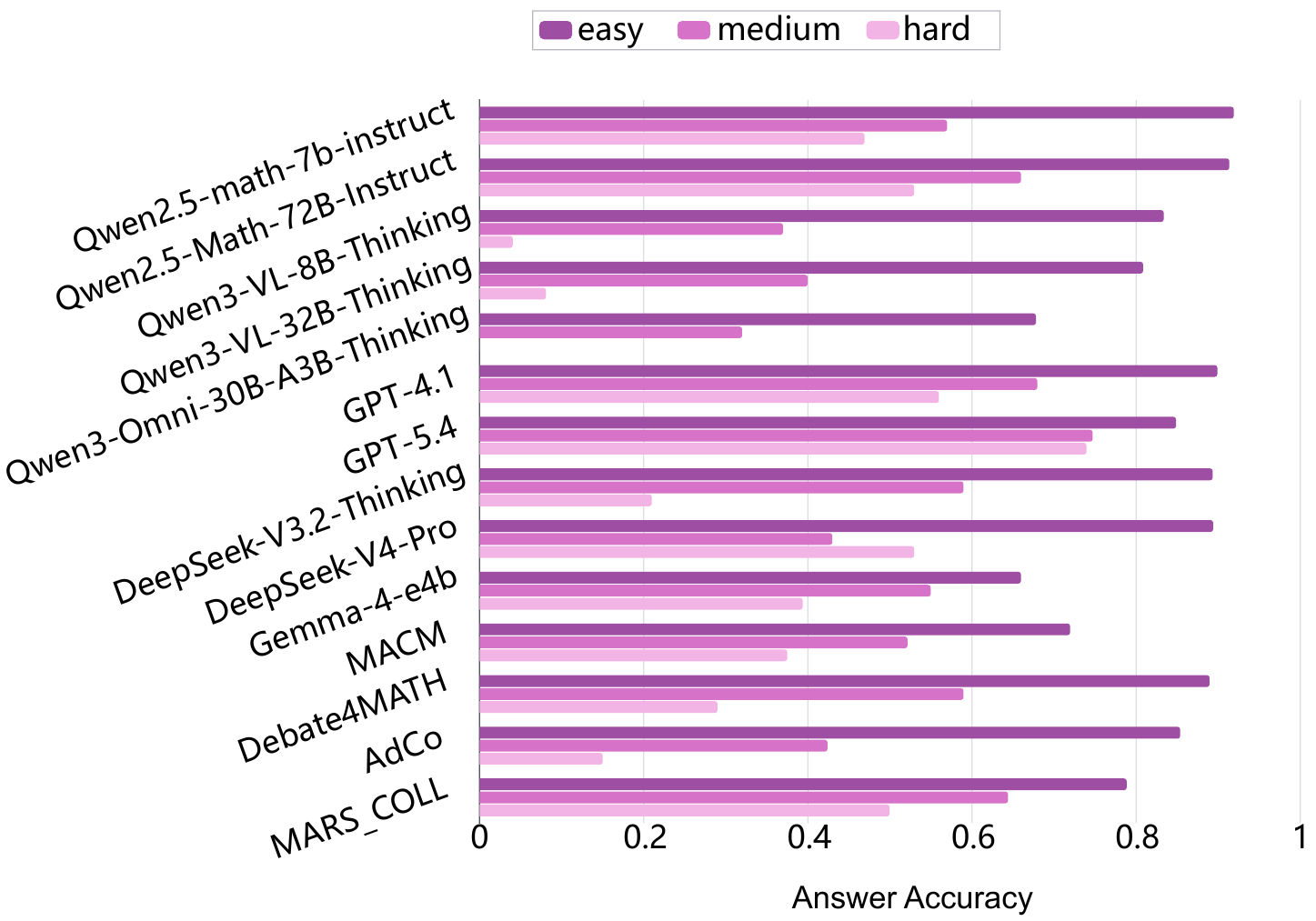}
  \caption{Performance across Difficulty Levels.}
  % \Description{Mathematical reasoning ability shows a downward trend as difficulty increases.}
  \label{fig:Performance across Difficulty Levels}
\end{figure}
Figure~\ref{fig:Performance across Difficulty Levels} illustrates the answer accuracy of all evaluated systems across three difficulty levels. Most models exhibited the expected monotonic trend: accuracy declined as task difficulty increased from easy to medium and then to hard, aligning with the inherent difficulty gradient of mathematical reasoning tasks.

Strong LLM baselines, including GPT-4.1 and GPT-5.4, maintained high accuracy on easy samples and competitive performance on medium-difficulty tasks; however, they still experienced a significant drop in performance when handling hard problems. DeepSeek-V4-Pro is a notable exception, achieving higher accuracy on the hard subset than on the medium subset. This behavior may reflect its stronger capabilities in complex mathematical reasoning or differences in the mathematical domains represented across difficulty levels, suggesting that model performance is not determined solely by problem difficulty.

When comparing single LLMs with multi-agent frameworks, MASs such as Debate4MATH and MARS\_COLL achieved accuracy levels comparable to top-tier single LLMs on easy tasks. However, when faced with medium and hard inputs, multi-agent frameworks failed to outperform powerful standalone LLMs, generally showing lower performance on hard samples. This indicates that current multi-agent collaboration yields limited benefits for challenging mathematical reasoning tasks and fails to effectively mitigate the performance degradation caused by high task complexity.

Overall, these results corroborate our earlier findings: multi-agent collaboration does not universally enhance mathematical reasoning performance, and the benefits of current multi-agent collaboration diminish under the more challenging subsets of MAMath-Bench. While MASs perform comparably to single models on easy problems, they fail to deliver substantial performance gains in difficult mathematical reasoning tasks involving complex, multi-step logic.

\subsection{Process Reliability of Collaborative Reasoning}
\subsubsection{Final-Answer Accuracy vs. Milestone Accuracy}
\begin{figure}[t]
  \centering
  \includegraphics[width=0.9\columnwidth]{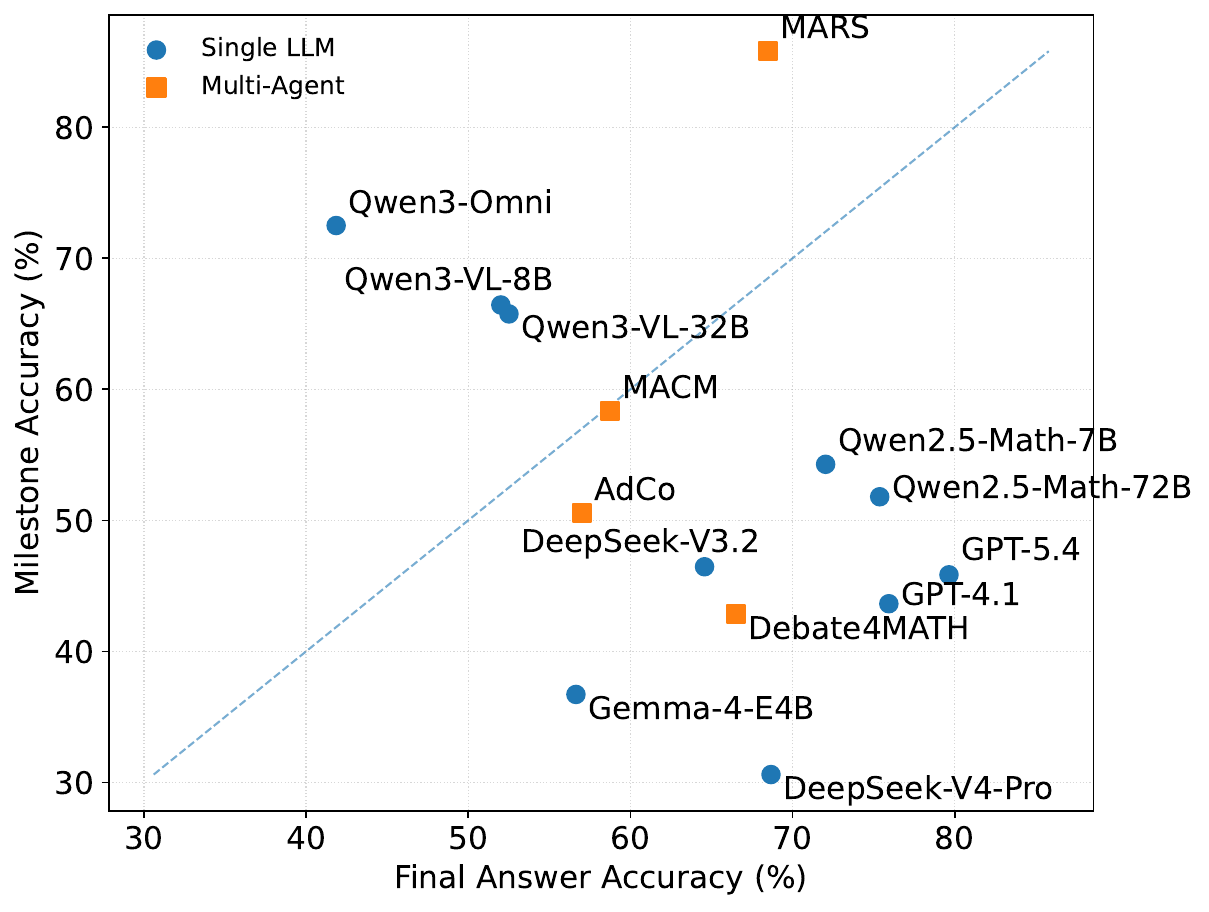}
  \caption{Relationship between FAA and MA on the text-only subset.}
  % \Description{This figure illustrates the capabilities across different dimensions of the challenging problems, as characterized by process-oriented and outcome-oriented indicators.}
  \label{fig:Final-Answer Accuracy vs. Milestone Accuracy}
\end{figure}

To investigate whether the correctness of the final answer reflects the reliability of the underlying reasoning process, Figure~\ref{fig:Final-Answer Accuracy vs. Milestone Accuracy} compares the FAA and MA across all tested systems.
The results reveal a clear mismatch between outcome-level correctness and process-level reasoning quality. Systems with high FAA do not necessarily exhibit high milestone accuracy. Despite achieving high FAA, GPT-5.4 and GPT-4.1 exhibit relatively limited milestone coverage. Conversely, Qwen3-Omni achieves higher milestone accuracy despite its weaker final-answer performance. These findings suggest that strong problem-solving capabilities do not necessarily translate into comprehensive intermediate reasoning, highlighting distinct strengths and limitations across models.

Different multi-agent frameworks exhibit vastly different relationships between answers and processes. MARS\_COLL is positioned in the upper region of the plot, far exceeding FAA. This pattern is consistent with MARS\_COLL's iterative review mechanism, which may contribute to broader intermediate reasoning coverage. In contrast, MACM lies near the diagonal, reflecting a relatively balanced relationship between outcome correctness and process reliability. Debate4MATH shows the opposite trend: its FAA is substantially higher than that of MACM, suggesting that multi-agent debate can yield correct conclusions without consistently retaining all reference reasoning milestones. AdCo falls between these extremes, exhibiting moderate performance on both dimensions. 

These differences demonstrate that multi-agent frameworks should not be evaluated solely based on FAA, as distinct collaboration mechanisms give rise to diverse reasoning behaviors.

\subsection{Adversarial Reasoning Robustness}

\begin{figure}[t]
  \centering
  \includegraphics[width=\columnwidth]{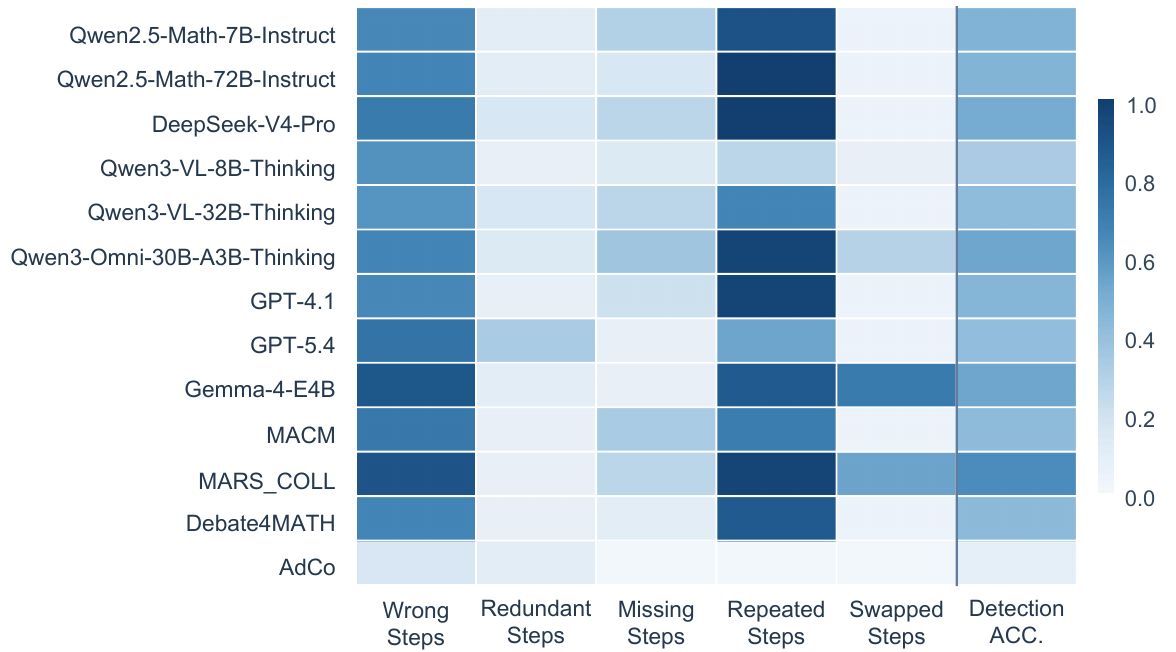}
  \caption{Detection accuracy across five types of adversarial reasoning perturbations. DeepSeek-V3.2-Thinking is excluded because its API was unavailable during the adversarial evaluation.}
  % \Description{Detection accuracy across five types of adversarial reasoning perturbations.}
  \label{fig:Complete Adversarial Results}
\end{figure}

Figure~\ref{fig:Complete Adversarial Results} presents a more granular comparison across five types of adversarial reasoning flaws: wrong steps, redundant steps, missing steps, repeated steps, and swapped steps. Detection performance varies significantly across these error types.

The results for repeated steps and wrong steps are particularly striking. Several systems achieved extremely high detection accuracy for these types of perturbations. In contrast, detecting redundant steps proved consistently difficult. All tested systems showed detection accuracy below approximately 33\%, with many models hovering near 10\%. Regarding missing steps, even powerful systems achieved only moderate accuracy. In the detection of swapped steps, Qwen3-VL-32B demonstrated the strongest performance, while most other models showed accuracy close to zero. This is likely because detecting structural ordering errors requires explicitly modeling the dependencies between intermediate reasoning steps, rather than simply verifying the correctness of individual statements.

Overall, these results highlight two broad categories of reasoning flaws:
\begin{itemize}[
    itemsep=0pt,
    parsep=0pt,
    partopsep=0pt
]
    \item Locally observable errors, such as explicit repetitions or obviously incorrect steps, which are relatively easy to detect;
    \item Global structural errors, such as redundancies, omissions, or sequencing errors, which require an understanding of dependencies across the entire reasoning process.
\end{itemize}

\section{Limitations and Future Directions}

Despite the comprehensive evaluation conducted by MAMath-Bench, several limitations remain.

\textbf{The evaluated multi-agent frameworks adopt different interaction protocols.} Therefore, performance differences across debate, condition-oriented collaboration, and cooperative-competitive mechanisms cannot be attributed solely to their collaboration strategies. Future work will adopt standardized agent configurations and controlled ablations to isolate the effects of individual collaboration mechanisms better.

\textbf{The current benchmark does not yet fully capture the diversity and complexity of collaborative mathematical reasoning.} Although MAMath-Bench encompasses various mathematical categories, difficulty levels, modalities, and adversarial errors, further evaluation is required regarding longer reasoning chains, richer cross-modal dependencies, and agent-level interactions. Future extensions will not only expand the benchmark's content but also introduce metrics for interaction-level aspects such as agent contributions, communication efficiency, consensus formation, and error propagation.

\textbf{Data contamination affects fairness to a certain extent.} Some problems in MAMath-Bench are adapted from publicly available mathematical reasoning datasets. We cannot completely rule out the possibility that some source problems were encountered during pretraining. This issue may affect absolute performance estimates. Nevertheless, MAMath-Bench focuses not only on final-answer correctness but also on process-level milestone alignment and adversarial reasoning-error detection, which require additional evaluation beyond reproducing known answers. Future versions will incorporate newly constructed and temporally controlled problems to mitigate potential contamination further.

\section{Conclusion}

We introduce MAMath-Bench, a benchmark designed to evaluate the multimodal mathematical reasoning capabilities of MASs across three complementary dimensions: solution correctness, process reliability, and robustness to adversarial reasoning-chain perturbations. By integrating multimodal mathematical problems, annotations of reasoning processes based on key steps, and adversarial reasoning perturbations, the benchmark enables an evaluation that goes beyond the traditional metric of final-answer accuracy. Experimental results demonstrate that multi-agent collaboration does not offer a universal advantage in terms of solution accuracy; rather, distinct collaboration mechanisms reshape the reasoning process in vastly different ways. Specifically, increasing the coverage of intermediate reasoning steps often introduces significant redundancy, whereas concise reasoning may compromise process completeness. Furthermore, even powerful systems are prone to structurally subtle reasoning errors, revealing a significant gap between answer correctness and reasoning reliability.
%%%%%%%%%%%%%%%%%%%%%%%%%%%%%%%%%%%%%%%%%%%%%%%%%%%%%%%%%%%%%%%%%%%%%%%%

%%% The acknowledgments section is defined using the "acks" environment
%%% (rather than an unnumbered section). The use of this environment 
%%% ensures the proper identification of the section in the article 
%%% metadata as well as the consistent spelling of the heading.

\section*{Acknowledgment}
This work was supported by the National Natural Science Foundation of China (Grant Nos. 62677055, 62477037), the China Postdoctoral Science Foundation (Grant Nos. 2024M752585, 2025T180425), the Key Research and Development Program of Shaanxi (Grant Nos. 2026CY-YBXM-854, 2024PT-ZCK-6), and the Youth Innovation Team of Shaanxi Universities "Multi-modal Data Mining and Fusion."

\bibliographystyle{ACM-Reference-Format} 
\bibliography{sample}

\appendix

\section{Additional Experimental Details}
\label{app:experimental_details}
This appendix provides additional implementation details to ensure the reproducibility of our experiments. 

\subsection{Model Inference Configuration}
\label{app:model_configuration}
To ensure a fair and reproducible evaluation, we follow the default inference configurations recommended for each evaluated model. Specifically, generation parameters such as temperature and maximum output tokens are kept unchanged from their official settings rather than manually optimized for individual models. This avoids introducing model-specific prompt or hyperparameter tuning and provides a more realistic comparison of their out-of-the-box reasoning capabilities.

Table~\ref{tab:inference_config} summarizes the inference configurations used in our experiments.

\begin{table}[h]
\centering
\setlength{\tabcolsep}{4pt}
\caption{Inference configuration of evaluated models.}
\label{tab:inference_config}
\begin{tabular}{lccc}
\toprule
Model & Backend & Temperature & Max Tokens\\
\midrule
\multicolumn{4}{l}{\textit{Single LLM}} \\
\midrule
Qwen2.5-Math-7B      & Local & Default & Default \\
Qwen2.5-Math-72B     & API   & Default & Default \\
Qwen3-VL-8B          & Local & Default & Default \\
Qwen3-VL-32B         & Local & Default & Default \\
Qwen3-Omni-30B       & Local & Default & Default \\
GPT-4.1              & API   & Default & Default \\
GPT-5.4              & API   & Default & Default \\
DeepSeek-v3.2        & API   & Default & Default \\
DeepSeek-V4-Pro      & API   & Default & Default \\
Gemma-4-E4B          & Local & Default & Default \\
\midrule
\multicolumn{4}{l}{\textit{Multi-Agent Framework}} \\
\midrule
Debate4MATH          & API   & Default & Default \\
AdCo                 & API   & Default & Default \\
MARS\_COLL           & API   & Default & Default \\
MACM                 & API   & Default & Default \\
\bottomrule
\end{tabular}
\end{table}

\subsection{Prompt Templates}
\label{app:prompt}

\subsubsection{Positive Sample Prompts}
To ensure a fair comparison, all evaluated models receive standardized mathematical reasoning instructions. The same task instruction is applied across different models, while the input modality is adjusted according to the corresponding problem setting.

The system prompt is:

\begin{quote}
You are a helpful assistant capable of math reasoning. Please reason step by step, and put your final answer within \textbackslash boxed\{\}.
\end{quote}

For multimodal problems, the prompt construction module preserves the original modality information and converts each instance into a unified message representation. Text-only inputs are directly inserted into the user message, while image and audio inputs are attached as corresponding multimodal content blocks.

\subsubsection{Negative Sample Prompts}

For robustness evaluation, we design task-specific prompts for identifying reasoning defects in corrupted solution chains. Unlike standard mathematical solving prompts, these instructions explicitly require models to analyze existing reasoning processes rather than solve the problem from scratch. The prompts constrain the output format to ensure consistent evaluation of error detection performance.

\begin{itemize}[
    itemsep=0pt,
    parsep=0pt,
    partopsep=0pt
]
    \item \textbf{Wrong Step:} The model is required to locate the first incorrect reasoning step caused by an invalid calculation or logical mistake. The expected output is the index of the incorrect step.

    \item \textbf{Redundant Step:} The model is instructed to identify unnecessary reasoning steps that do not contribute to solving the problem. The expected output is a list of redundant step indices.

    \item \textbf{Repeated Step:} The model is required to detect duplicated reasoning steps appearing in the solution chain. The expected output is the index of the repeated step.

    \item \textbf{Missing Step:} The model is asked to identify omitted reasoning units that are necessary for completing the solution process. The expected output includes the missing step and its corresponding position.

    \item \textbf{Swapped Step:} The model is required to detect two reasoning steps whose order has been incorrectly exchanged. The expected output is the pair of swapped step indices.
\end{itemize}
\subsection{Computational Cost and Token Statistics}
\label{app:token_statistics}

To analyze the computational overhead associated with different reasoning paradigms, we record the average token consumption and inference duration for each evaluated system. Specifically, we measure the total number of generated and consumed tokens during inference, as well as the overall execution time required to process each problem instance. These statistics provide a complementary perspective on the efficiency of different reasoning approaches beyond their task performance.

Table~\ref{tab:token_stats} summarizes the average token usage and inference duration of evaluated models. Single LLMs generally complete inference through a single generation process and therefore maintain relatively moderate computational costs. Results show that DeepSeek-v3.2 and the GPT series consume the fewest tokens. Multimodal models tend to introduce additional computational overhead due to the processing of visual and audio inputs, resulting in higher token consumption and longer inference time compared with text-only models.

\begin{table}[htbp]
\centering
\setlength{\tabcolsep}{4pt}
\caption{Average token consumption and inference duration of evaluated models.}
\label{tab:token_stats}
\begin{tabular}{lcc}
\toprule
Model & Avg. Total Tokens & Avg. Duration (s)\\
\midrule
Qwen2.5-Math-7B              & 739.82 & 10.87 \\
Qwen2.5-Math-72B             & 716.21 & 20.91 \\
Qwen3-VL-8B                  & 3079.68 & 52.98 \\
Qwen3-VL-32B                 & 3031.06 & 96.83 \\
Qwen3-Omni-30B               & 2301.58 & 53.59 \\
GPT-4.1                       & 985.52 & 6.78 \\
GPT-5.4                       & 915.12 & 9.62 \\
DeepSeek-v3.2                 & 906.60 & 59.31 \\
DeepSeek-V4-Pro               & 1320.52 & 11.01 \\
Gemma-4-E4B                   & 1671.07 & 153.43 \\

\bottomrule
\end{tabular}
\end{table}

% \midrule
% \multicolumn{3}{l}{\textit{Multi-Agent Framework}} \\
% \midrule
% Debate4MATH                   & 62392.72 & 84.47 \\
% AdCo                          & -- & -- \\
% MARS\_COLL                    & -- & 100.88 \\
% MACM                          & 11925.98 & 58.09 \\

\end{document}